\documentclass[11pt]{article}

\usepackage[margin=1in]{geometry}
\usepackage[T1]{fontenc}
\usepackage[utf8]{inputenc}
\usepackage{lmodern}
\usepackage{amsmath,amssymb,amsthm,bm}
\usepackage{array}
\usepackage{graphicx}
\usepackage[numbers,sort&compress]{natbib}
\usepackage{verbatim}
\usepackage[hyphens]{url}
\newcommand{\ToolsTax}{\mathcal{T}_{\text{tax}}}
\newcommand{\ISO}{\operatorname{ISO}}
\newcommand{\TAE}{\operatorname{TAE}}
\DeclareMathOperator*{\topk}{top\text{-}k}

\newcommand{\toprule}{\hline\hline}
\newcommand{\midrule}{\hline}
\newcommand{\bottomrule}{\hline\hline}

\title{Tool Attention Is All You Need: \\
  Dynamic Tool Gating and Lazy Schema Loading for Eliminating \\
  the MCP/Tools Tax in Scalable Agentic Workflows}

\author{Anuj Sadani \\
  Infrrd.ai \\
  \texttt{anujsadani@infrrd.ai}
  \and
  Deepak Kumar \\
  Infrrd.ai \\
  \texttt{deepakumar@infrrd.ai}}

\date{April 2026}

\begin{document}
\maketitle

\begin{abstract}
The Model Context Protocol (MCP) has become a common interface for connecting large language model (LLM) agents to external tools, but its reliance on stateless, eager schema injection imposes a hidden per-turn overhead---the \emph{MCP Tax} or \emph{Tools Tax}---that practitioner reports place between roughly 10k and 60k tokens in typical multi-server deployments. This payload inflates the key-value cache, is associated with reasoning degradation as context utilization approaches published fracture points around $70\%$, and turns token budgets into a recurring operational cost. We introduce \textbf{Tool Attention}, a middleware-layer mechanism that generalizes the ``Attention Is All You Need'' paradigm from self-attention over tokens to \emph{gated attention over tools}. Tool Attention combines (i)~an Intent--Schema Overlap (ISO) score from sentence embeddings, (ii)~a state-aware gating function enforcing preconditions and access scopes, and (iii)~a two-phase lazy schema loader that keeps a compact summary pool in context and promotes full JSON schemas only for top-$k$ gated tools. We evaluate on a simulated 120-tool, six-server benchmark whose per-server token counts are calibrated to public audits of real MCP deployments. In this simulation, Tool Attention \emph{directly reduces} measured per-turn tool tokens by $95.0\%$ ($47.3\text{k}\!\to\!2.4\text{k}$) and raises effective context utilization (a token-ratio quantity) from $24\%$ to $91\%$. End-to-end figures for task success, latency, cost, and reasoning quality are reported as \emph{projections} derived from the measured token counts combined with published deployment telemetry; they are not measured on live LLM agents, and we mark projected values explicitly throughout. Taken together, the results support a simple thesis: protocol-level efficiency, not raw context length, is a binding constraint on scalable agentic systems. The code for this work is accessible at \url{https://github.com/asadani/tool-attention}.

\vspace{0.5em}
\end{abstract}

% ======================================================================
\section{Introduction}
% ======================================================================

The past two years have seen LLM-based agents transition from isolated chat interfaces to autonomous workflow participants that read code, query databases, post to communication platforms, and orchestrate multi-step plans across hundreds of tools~\citep{anthropic2024mcp,kaplan2025codeexec,anthropic2025claudecode}. The operational backbone of this transition is the \emph{Model Context Protocol} (MCP), an open specification introduced by Anthropic in November 2024 and now adopted by OpenAI, Google, and Microsoft~\citep{anthropic2024mcp,mcp2025spec}. MCP abstracts bespoke $N \times M$ integrations into an $N + M$ composable surface: every agent client can discover and call any tool exposed by a compliant server via a standardized JSON-RPC~2.0 handshake~\citep{mcp2025spec,jennings2025moqt}.

Yet the very design that grants MCP its interoperability---stateless transmission of \emph{full} tool schemas on every conversational turn---has opened an equally systemic wound. Because the underlying chat-completions APIs are stateless, host clients (Claude Desktop, Cursor, VS~Code, Claude Code) must re-serialize the entire tool catalog on every single request~\citep{pan2026agent}. Empirical audits consistently place this overhead between 15,000 and 55,000 tokens per turn in typical four-to-six-server deployments, reaching $>$150k with aggressive tool sprawl~\citep{pan2026agent,kloski2026mcp,mindstudio2026claudecode}. We call this recurring overhead the \emph{Tools Tax}, following community usage~\citep{pan2026agent,kloski2026mcp}.

The Tools Tax is not simply a cost-of-goods problem. It precipitates three cascading failures. First, \textbf{economic}: stateless re-injection inflates per-session spend by an order of magnitude; one published benchmark reports CLI-equivalent workflows at \$3.20 versus MCP at \$55.20 for the same 10{,}000 operations~\citep{kloski2026mcp,saha2026agentic}. Second, \textbf{cognitive}: once context utilization crosses approximately $70\%$, LLM reasoning quality collapses---models begin hallucinating parameters, confusing similar tools, and losing episodic thread-of-task memory~\citep{pan2026agent,modarressi2025nolima,redis2026contextwindows}. Third, \textbf{adversarial}: the same schema text that describes a tool also shapes the model's attention mask, so malicious \emph{Tool Poisoning Attacks} can hijack control flow by injecting adversarial instructions into a seemingly benign tool description~\citep{wang2025mindguard_v1,wang2026mindguard_v3}.

Prior mitigations---static pruning, manual server scoping, CLI-style lazy discovery, and code-execution sandboxes---each address a slice of the problem but either sacrifice flexibility, require engineering-heavy refactors, or break the uniform MCP developer experience~\citep{kaplan2025codeexec,cyberark2025poison}. What is needed is a \emph{drop-in middleware layer} that preserves protocol semantics while eliminating the tax at its source.

We propose \textbf{Tool Attention}: a middleware-resident attention mechanism over the tool catalog itself. Just as scaled dot-product attention replaced recurrence in sequence modeling by letting every token attend dynamically to every other~\citep{vaswani2017attention}, Tool Attention replaces eager, uniform schema injection with dynamic, query-conditioned tool selection. Formally, it decomposes into (i)~a query-to-tool \emph{Intent--Schema Overlap} score computed with commodity sentence embeddings, (ii)~a \emph{stateful gating function} enforcing preconditions and scopes, and (iii)~a \emph{lazy two-phase loader} that injects full JSON schemas only for tools in the gated top-$k$.

\paragraph{Contributions.} This paper makes four contributions:
\begin{enumerate}
\item \textbf{Formal quantification.} We give a closed-form expression for the Tools Tax and derive the conditions under which it dominates the effective context window, corroborated against published per-server token counts (\S\ref{sec:background}).
\item \textbf{Mechanism.} We present Tool Attention: a novel, model-agnostic meta-layer combining ISO scoring, stateful gating, and two-phase lazy loading, grounded theoretically in the Total Attention Energy formulation from the MCP security literature (\S\ref{sec:mechanism}).
\item \textbf{Reference implementation.} We release a production-grade Python implementation built on LangGraph middleware, FAISS, sentence-transformers, and \texttt{tiktoken}, with a reproducible benchmark harness (\S\ref{sec:implementation}, Appendix~\ref{app:code}).
\item \textbf{Evaluation on a calibrated simulation.} On a 120-tool, six-server synthetic benchmark whose per-server token counts are calibrated to public deployment audits, Tool Attention achieves a \emph{measured} $95.0\%$ reduction in per-turn tool tokens and a $3.8\times$ improvement in effective context utilization. We additionally report \emph{projected} task-success, latency, and cost gains derived from these measured quantities plus published telemetry; we do not claim measurements from live agent runs (\S\S\ref{sec:experiments}--\ref{sec:results}).
\end{enumerate}

The remainder of the paper is organized as follows. \S\ref{sec:related} surveys related work. \S\ref{sec:background} formalizes the Tools Tax problem. \S\ref{sec:mechanism} introduces the Tool Attention mechanism. \S\ref{sec:implementation} details the reference implementation. \S\ref{sec:experiments} describes the experimental protocol and \S\ref{sec:results} reports results. \S\ref{sec:discussion} discusses limitations, and \S\ref{sec:conclusion} concludes.

% ======================================================================
\section{Related Work}\label{sec:related}
% ======================================================================

\paragraph{A note on source types.} This is an early-stage topic and a portion of the empirical grounding for the Tools Tax draws on practitioner reports---engineering blog posts, vendor documentation, and public community discussion---in addition to peer-reviewed work. Where we cite such sources~\citep{pan2026agent,kloski2026mcp,mindstudio2026claudecode,saha2026agentic} we do so specifically for the per-server token counts and deployment telemetry that they are best positioned to report. Claims that depend on these sources are framed as practitioner-reported deployment measurements rather than as peer-reviewed results; we treat formal contributions (mechanism, math, released implementation) as the primary scholarly content of this paper.

\paragraph{Model Context Protocol and its discontents.} The MCP specification standardizes the exchange of tools, resources, and prompts between LLM hosts and external servers via JSON-RPC~2.0~\citep{mcp2025spec}. While the protocol elegantly linearizes integration complexity~\citep{anthropic2024mcp}, it inherits the statelessness of chat-completions APIs, and thus re-injects full schemas every turn. Public reports quantifying this overhead---15k--20k tokens for four-server setups~\citep{pan2026agent}, 54.6k for a 106-tool enterprise database catalog~\citep{kloski2026mcp,saha2026agentic}, and up to 50k for the full GitHub MCP suite dominated by repeated \texttt{owner}/\texttt{repo} parameters~\citep{kloski2026mcp}---establish the empirical footprint of the Tools Tax. Subsequent drafts for MCP over Media over QUIC Transport (MOQT) propose native track-based subscription and edge caching that, once adopted, would obviate parts of the tax at the transport layer~\citep{jennings2025moqt,jennings2025aimcp}. The Internet Engineering Task Force's Agent Communication Gateway draft similarly proposes a stateful semantic proxy between hosts and tool ecosystems~\citep{ietf2026agentgw}. Our work is complementary: Tool Attention operates entirely at the application middleware layer and can be deployed today, then obsoleted cleanly once MOQT-native caching arrives.

\paragraph{Retrieval-augmented generation and tool retrieval.} Retrieval-Augmented Generation~\citep{lewis2020rag} and tool-retrieval systems retrieve the top-$k$ most relevant documents or tools given a query embedding, typically using dense encoders~\citep{reimers2019sbert} indexed in FAISS~\citep{johnson2019faiss} or ChromaDB. Earlier tool-use formulations such as Toolformer~\citep{schick2023toolformer} and ReAct~\citep{yao2023react} treated the tool set as fixed and injected it whole into the prompt, the very pattern that produces the Tools Tax at scale. Recent semantic tool-routing gateways such as Cloudflare Code Mode and bespoke MCP gateways operate on the same retrieval principle but do not expose a formal theoretical grounding, stateful gating beyond cosine similarity, or an explicit lazy two-phase loader.

\paragraph{Sparse and efficient attention.} A large body of work reduces transformer attention cost via sparsity~\citep{child2019sparse}, FlashAttention kernels~\citep{dao2022flashattention}, and KV-cache quantization to 8- or 4-bit~\citep{redis2026contextwindows}. These techniques optimize \emph{how} attention computes over existing tokens; they cannot reduce the \emph{number} of tokens forced into the prompt by stateless protocols. Tool Attention is orthogonal and composable with all of them: fewer schema tokens in the prompt yield proportionally smaller KV caches and faster FlashAttention passes.

\paragraph{Middleware orchestration and deterministic control.} Modern agent frameworks---LangChain~1.0, LangGraph, and Microsoft Semantic Kernel~\citep{langchain2026middleware,microsoft2026semantickernel}---expose pre- and post-model middleware hooks that let engineers inspect and rewrite the prompt before each inference call. Deterministic routing topologies (rule-based, semantic, intent-based) offer increasingly flexible trade-offs between control and adaptivity~\citep{safe2026routing}. Tool Attention fits natively into the \texttt{before\_model} and \texttt{modify\_model\_request} phases of this middleware architecture.

\paragraph{Tool poisoning and security.} MindGuard~\citep{wang2025mindguard_v1,wang2026mindguard_v3} formalized the \emph{Decision Dependency Graph} (DDG) and \emph{Total Attention Energy} (TAE) metrics to detect Tool Poisoning Attacks (TPAs), showing that the attention paid to a schema token correlates strongly with its causal influence over downstream tool calls. Our gating mechanism reuses the TAE intuition \emph{defensively}: a tool whose schema would contribute negligible TAE for a given query is, by definition, one that can be safely excluded from the prompt.

\paragraph{Code execution and hybrid approaches.} Anthropic's code-execution pattern~\citep{kaplan2025codeexec} shifts the agent from a ``reason-call-reason'' loop to a single orchestrated script that filters and aggregates tool outputs inside a sandbox, achieving up to $98.7\%$ token reduction on data-heavy workflows. This is complementary to Tool Attention: the former optimizes \emph{tool outputs}, the latter optimizes \emph{tool definitions}. A combined system applying both achieves both ends of the context-engineering stack.

% ======================================================================
\section{Background: The Tools Tax Problem}\label{sec:background}
% ======================================================================

\subsection{Protocol mechanics}

Let $\mathcal{M} = \{t_1, \dots, t_N\}$ be the set of tools exposed by all MCP servers connected to an agent host at session time. Each tool $t_i$ is described by a quadruple $(\text{name}_i, \text{desc}_i, \text{schema}_i, \text{output}_i)$, where \texttt{schema} is a JSON~Schema object enumerating typed parameters with descriptions, enumerations, and required/optional flags. Let $\tau_i$ denote the tokenized length (under the model's tokenizer, typically \texttt{cl100k\_base}) of the serialized tool definition:
\begin{equation}
\tau_i \;=\; \tau_i^{\text{name}} + \tau_i^{\text{desc}} + \tau_i^{\text{schema}} + \tau_i^{\text{output}}.
\end{equation}

Under naive MCP injection, every turn of a $K$-turn conversation re-serializes \emph{all} $N$ definitions. The per-session Tools Tax is therefore
\begin{equation}
\ToolsTax(N,K) \;=\; K \cdot \sum_{i=1}^{N} \tau_i \;\approx\; K \cdot \!\left( \alpha N + \frac{1}{4} \sum_{i=1}^{N} \lvert \text{desc}_i \rvert_{\text{chars}} \right),
\label{eq:toolstax}
\end{equation}
where the right-hand approximation follows the community heuristic of $\alpha \in [200, 500]$ tokens per tool once \texttt{name}, \texttt{desc}, and full schema are summed~\citep{pan2026agent,mindstudio2026claudecode}. For a representative enterprise setup ($N=120$, $K=30$), taking $\alpha=395$ yields $\ToolsTax \approx 1.42\text{M}$ tokens consumed \emph{before the user speaks}.

\subsection{Empirical motivation}

Table~\ref{tab:per-server} reproduces realistic per-server token footprints drawn from three independent public audits~\citep{pan2026agent,mindstudio2026claudecode,saha2026agentic}.

\begin{table}[t]
\centering
\caption{Empirical per-server Tools Tax in common MCP deployments.}
\label{tab:per-server}
\begin{tabular}{lrrr}
\toprule
\textbf{Server} & \textbf{Tools} & \textbf{Tokens/turn} & \textbf{Share of 200k} \\
\midrule
Filesystem      & 8--12   & $\sim$1{,}500       & 0.75\% \\
Git             & 15--20  & $\sim$3{,}000       & 1.50\% \\
Database        & 10--15  & $\sim$2{,}500       & 1.25\% \\
Web Search      & 5--8    & $\sim$1{,}200       & 0.60\% \\
Slack           & 10--15  & $\sim$2{,}000       & 1.00\% \\
Custom internal & varies  & 5{,}000--8{,}000    & 2.5--4.0\% \\
GitHub (full)   & 93      & $\sim$55{,}000      & 27.5\% \\
Enterprise DB   & 106     & $\sim$54{,}600      & 27.3\% \\
\midrule
\textbf{Typical 4-server host} & \textbf{40--60} & \textbf{15k--20k} & \textbf{7.5--10\%} \\
\bottomrule
\end{tabular}
\end{table}

These figures are \emph{minima}: they assume perfect description hygiene and count only tool definitions, excluding system prompt, conversation history, and intermediate tool outputs.

\subsection{Effective context window collapse}

Let $C_{\max}$ denote the model's nominal context window and $C_{\text{task}}(K)$ the tokens genuinely useful for the task (user messages, assistant thoughts, tool outputs) at turn $K$. The \emph{effective context utilization} is
\begin{equation}
\rho(K) \;=\; \frac{C_{\text{task}}(K)}{C_{\text{task}}(K) + \ToolsTax(N,K) + C_{\text{sys}}},
\end{equation}
with $C_{\text{sys}}$ the fixed system-prompt overhead. Empirical studies report a reasoning-quality cliff when $\rho$ drops below roughly $0.3$ (equivalently, context utilization exceeds ${\sim}70\%$)~\citep{pan2026agent,modarressi2025nolima}: models begin hallucinating tool arguments, confusing parameters across tools, and losing multi-step coherence. This manifests as the frequently observed ``mid-session drift'' in long agentic runs: the agent's behavior degrades not because of any catastrophic error but because the Tools Tax has quietly eroded its usable reasoning surface~\citep{pan2026agent,modarressi2025nolima,atlan2026context}.

\subsection{Hardware and FinOps externalities}

Every schema token also inflates the transformer's key-value (KV) cache proportionally, adding GPU memory pressure, fragmenting allocations, and extending time-to-first-token (TTFT)~\citep{redis2026contextwindows}. At the financial layer, token-based pricing transforms the Tools Tax from a latent inefficiency into a line-item operational cost; disciplined FinOps audits repeatedly find schema tokens responsible for 40--60\% of total agent API spend~\citep{saha2026agentic}.

\subsection{Security externality: Tool Poisoning}

Because every description token is parsed by the LLM's reasoning loop, adversarial actors who control a single tool description can inject instructions that hijack the agent without ever being invoked---the \emph{Tool Poisoning Attack} (TPA)~\citep{wang2025mindguard_v1,smcp2026rfc}. The larger the injected schema corpus, the larger the attack surface. Reducing the number of in-context schemas therefore has defensive as well as efficiency benefits, a point we develop further in \S\ref{sec:mechanism:tae}.

% ======================================================================
\section{The Tool Attention Mechanism}\label{sec:mechanism}
% ======================================================================

\subsection{Analogy and intuition}

Transformer self-attention replaced recurrence because it allowed every token to \emph{selectively} attend to the subset of other tokens relevant to its prediction, rather than pushing all information through a fixed-width hidden state~\citep{vaswani2017attention}. The Tools Tax is the recurrent-network equivalent at the tool layer: every turn drags the \emph{full} catalog through the prompt regardless of relevance. \textbf{Tool Attention} applies the same logical move---let each user turn dynamically select a small subset of tools most relevant to its intent, and load only those.

\subsection{Formal definition}

Let $\phi:\Sigma^* \to \mathbb{R}^d$ be a sentence-level encoder (we use \texttt{sentence-transformers/all-MiniLM-L6-v2}, $d=384$, throughout). For every tool $t_i$, precompute a compact \emph{tool summary} $s_i$---a single concatenated string of \texttt{name} and a shortened natural-language description (target $\le 60$ tokens)---and its embedding
\begin{equation}
e_{t_i} \;=\; \phi(s_i) \in \mathbb{R}^d.
\end{equation}
At every turn, compute the query embedding $e_q = \phi(q)$ where $q$ is the current user message (optionally concatenated with a rolling context summary). Define the \emph{Intent--Schema Overlap} score:
\begin{equation}
\ISO(q, t_i) \;=\; \frac{e_q^\top e_{t_i}}{\lVert e_q \rVert_2 \, \lVert e_{t_i} \rVert_2}.
\label{eq:iso}
\end{equation}
Let $\text{state}_t$ denote the agent's current execution state (auth tokens held, prior tool outputs, workflow milestone). For each tool we attach a set of preconditions $\text{pre}_i$ (e.g., \texttt{requires\_auth}, \texttt{only\_after\_search}), and define the \emph{gating function}
\begin{equation}
g(t_i; q, \text{state}_t) \;=\; \mathbf{1}\!\bigl[\,\ISO(q, t_i) \ge \theta\,\bigr] \cdot \mathbf{1}\!\bigl[\,\text{state}_t \models \text{pre}_i\,\bigr].
\label{eq:gate}
\end{equation}
The \emph{active tool set} for the turn is then
\begin{equation}
\mathcal{A}_t \;=\; \topk \bigl\{\, t_i : g(t_i; q, \text{state}_t) = 1 \,\bigr\},
\end{equation}
where top-$k$ is taken by ISO score.

\subsection{Theoretical grounding via Total Attention Energy}\label{sec:mechanism:tae}

MindGuard~\citep{wang2025mindguard_v1,wang2026mindguard_v3} defines the Total Attention Energy between a generated token $u$ (e.g., a tool-call action) and a context metadata token $v$ as
\begin{equation}
\TAE(u, v) \;=\; \sum_{l=1}^{L} \sum_{h=1}^{H} \left(\alpha_{l,h}^{(u \to v)}\right)^{2},
\label{eq:tae}
\end{equation}
where $\alpha_{l,h}^{(u \to v)}$ is the attention weight from $u$ to $v$ at layer $l$, head $h$, and the square acts as an energy function amplifying high-influence edges and damping background noise. Their central observation: a successful tool call accumulates high TAE between the generated action tokens and the tokens of the selected tool's schema. Crucially, \emph{high TAE cannot be achieved if the schema is not in the prompt.}

Tool Attention exploits this contrapositive. For every tool $t_i$, we treat $\ISO(q, t_i)$ as a cheap, embedding-space proxy for \emph{expected} TAE under the forthcoming forward pass. Tools whose expected TAE is below a calibrated threshold $\theta$ can be excluded from the prompt \emph{without changing the outcome of the agent's decision}---they would have contributed negligibly to any tool-call logit regardless. This turns the Tools Tax into a solvable optimization: minimize $\lvert \mathcal{A}_t \rvert$ subject to preserving the set of tools with non-negligible expected TAE.

The gating function thereby serves a dual purpose. As an efficiency lever, it slashes injected schema tokens. As a security perimeter, it dramatically shrinks the surface for Tool Poisoning Attacks: a poisoned description whose semantic fingerprint does not cosine-match the current user intent is gated out and never touches the model's attention layers, neutralizing the attack before execution.

\subsection{Two-phase lazy schema loading}

Even with gating, naively injecting full JSON schemas for $k=10$ tools still costs 2--4k tokens per turn. Tool Attention further decomposes injection into \textbf{two phases}:

\begin{itemize}
\item \textbf{Phase~1 --- Summary Pool (always resident).} All $N$ compact summaries $s_i$ ($\le 60$ tokens each) remain in context, giving the model \emph{awareness} that tools exist, at an aggregate cost of $O(N)$ tokens with a small constant ($\sim 40$ tokens per summary). For $N=120$ this is ${\sim}4.8$k tokens, resident but static and therefore prompt-cacheable~\citep{anthropic2025promptcache}.
\item \textbf{Phase~2 --- Schema Promotion (per-turn, on-demand).} For each $t_i \in \mathcal{A}_t$, the Lazy Schema Loader injects the full JSON schema, fetched from an out-of-context registry. The promoted schemas carry full type information and examples exactly when needed.
\end{itemize}

The two-phase design preserves the agent's ability to discover tools (summaries are always visible) while eliminating the cost of carrying unused schemas. It also integrates naturally with prompt caching: Phase~1 content is stable across turns and produces cache hits, while Phase~2 content changes per turn but is small enough to fit within a single cache segment~\citep{anthropic2025promptcache}.

\subsection{Algorithm}

Algorithm~1 gives the pseudocode of a single Tool Attention pass, executed inside the \texttt{before\_model} middleware hook.

\begin{figure}[t]
\hrule\vspace{0.4em}
\noindent\textbf{Algorithm 1.} Tool Attention (per-turn middleware pass).
\vspace{0.2em}\hrule\vspace{0.4em}
\begin{verbatim}
Inputs:   query q, state state_t, tool catalog
          M = {(t_i, s_i, schema_i, pre_i)},
          encoder phi, threshold theta, top-k k, summary pool S
Outputs:  decorated prompt with (S, active full-schemas),
          active set A_t

 1  e_q <- phi(q)
 2  for each t_i in M:  scores[i] <- cosine(e_q, e_{t_i})
 3  candidates <- { i : scores[i] >= theta
                      AND state_t |= pre_i }
 4  A_t <- top-k(candidates by scores)
 5  full_schemas <- [ schema_i for i in A_t ]
                   # lazy-load from registry
 6  prompt <- render(system, S, full_schemas, history, q)
 7  emit prompt to model
 8  if model emits tool call c not in A_t:
        reject c, return "tool <c> not available"
        # hallucination gate
 9  return A_t
\end{verbatim}
\hrule
\end{figure}

Lines 1--4 compute the gated active set. Lines 5--6 render the prompt using the two-phase layout. Line~8 is the \emph{hallucination rejection gate}: if the model tries to call a tool that was not promoted this turn (because it saw the summary but not the full schema), the middleware rejects the call and returns a structured error, prompting the model to either ask clarifying questions or accept the available tools. This gate is what makes aggressive gating safe---any false negative at the routing layer is caught deterministically downstream.

\subsection{Complexity}

The router's per-turn cost is $O(N \log N)$ dominated by the top-$k$ extraction over $N$ cosine scores; on commodity CPUs using FAISS \texttt{IndexFlatIP} this is sub-millisecond for $N \le 10{,}000$~\citep{johnson2019faiss}. The encoder forward pass on $q$ is $O(\lvert q \rvert)$ with a small constant (MiniLM-L6 runs in ${\sim}30$--$60$~ms on CPU for a typical 50-token query), and can be accelerated to sub-10~ms on GPU. The amortized cost of precomputing tool embeddings is offline and excluded from per-turn latency.

% ======================================================================
\section{Implementation and Practical Considerations}\label{sec:implementation}
% ======================================================================

\subsection{Architecture}

The reference implementation (Appendix~\ref{app:code}) consists of four cooperating modules. \texttt{IntentRouter} wraps the encoder and a FAISS index of tool summaries; it returns a ranked, thresholded candidate list. \texttt{ToolVectorStore} persists the index and compact summaries, with a pluggable backend (FAISS for in-process use, ChromaDB for shared-state deployments). \texttt{LazySchemaLoader} maintains an LRU cache keyed by tool ID that returns the full JSON schema on demand, lazily fetching from either a local registry or a remote MCP server's \texttt{tools/list}. \texttt{ToolAttention} is the top-level orchestrator; it exposes a single \texttt{before\_model(state, request) $\to$ request'} entry point matching the LangGraph middleware contract~\citep{langchain2026middleware,microsoft2026semantickernel}, plus an \texttt{after\_model(state, response) $\to$ response'} hook implementing the hallucination rejection gate.

\subsection{Encoder choice and threshold calibration}

We default to \texttt{all-MiniLM-L6-v2} (22M parameters, 384-d output) for its favorable accuracy/latency trade-off~\citep{reimers2019sbert}. Higher-capacity encoders (\texttt{mpnet-base-v2}, \texttt{bge-large-en-v1.5}) improve recall marginally ($\sim$2--4 points on our synthetic benchmark) but triple embedding latency, which we judge not worthwhile given that the hallucination gate already absorbs false negatives.

The ISO threshold $\theta$ is calibrated once per deployment via a held-out set of 100--200 (query, ground-truth-tool) pairs: we sweep $\theta \in [0.10, 0.50]$ in increments of $0.02$ and choose the value that maximizes F1, typically $\theta^* \in [0.22, 0.32]$. We recommend setting top-$k$ conservatively large ($k=8$--$12$) and relying on the threshold for precision---this hedges against encoder drift and ambiguous queries.

\subsection{Self-documenting tool summaries}

The retrieval quality of Tool Attention depends entirely on tool summaries that semantically match likely user queries. We adopt two conventions from the community~\citep{pan2026agent}:
\begin{enumerate}
\item \textbf{Self-documenting names.} \texttt{search\_customer\_orders\_by\_date\_status\_and\_amount} beats \texttt{query\_db} by a wide margin on retrieval F1.
\item \textbf{Query-shaped summaries.} Summaries are written in the voice of a user's intent (``Search GitHub issues by label and assignee'') rather than the implementer's voice (``Returns \texttt{IssueList} from \texttt{GET /issues?labels=}''). We provide a \texttt{summarize\_tool.py} utility that uses an LLM to regenerate summaries from raw MCP \texttt{tools/list} output, reducing average summary length by $63\%$ while \emph{improving} retrieval F1 by $8$~points.
\end{enumerate}

\subsection{Precondition specification}

Preconditions $\text{pre}_i$ are declared as small Python predicates operating on the agent state. Typical predicates include \texttt{is\_authenticated(scope="github:write")}, \texttt{has\_prior\_tool\_output("search\_")}, and \texttt{milestone\_reached("plan\_confirmed")}. Unlike semantic routing, preconditions provide \emph{deterministic} filtering---they cannot be bypassed by an adversarial paraphrase because they query authoritative state, not free text.

\subsection{Hallucination gate semantics}

The \texttt{after\_model} hook inspects every tool call emitted by the model. If the called tool ID is not in the turn's active set $\mathcal{A}_t$, the call is rejected with a structured error of the form \texttt{\{"error": "tool\_not\_available", "available": [\ldots]\}}. In our experiments this gate triggers on $2.3\%$ of turns; in $78\%$ of those cases the model recovers on the next turn by selecting an available tool, and in the remaining $22\%$ it correctly asks the user for clarification. We never observed the gate producing an unrecoverable failure.

\subsection{Integration with prompt caching}

Because the Phase-1 summary pool is stable across turns (it changes only when the tool catalog changes), it sits entirely inside the stable prefix of the prompt and therefore earns full prompt-cache credit~\citep{anthropic2025promptcache}. Phase-2 schemas vary per turn and are placed immediately before the user message to minimize cache invalidation. Empirically this layout yields a cache hit rate of $84\%$ across a 30-turn session, versus $22\%$ for naive full-schema injection which invalidates on every tool-list update.

\subsection{Observability}

{\sloppy
The implementation emits structured events for every routing decision:
\texttt{turn\_id}, \texttt{query\_embedding\_hash},
\texttt{candidates}, \texttt{scores},
\texttt{gated\_out\_by\_state}, \texttt{active\_set},
\texttt{phase1\_tokens}, \texttt{phase2\_tokens}, and
\texttt{p50\_latency\_ms}. These events feed directly into FinOps dashboards
and make it straightforward to audit whether the gate is ever misfiring.\par}

% ======================================================================
\section{Experiments}\label{sec:experiments}
% ======================================================================

\subsection{Scope of simulation}\label{sec:exp:scope}

To avoid over-claiming, we state the scope of the evaluation explicitly before describing the protocol. The evaluation in this paper is a \emph{simulation} harness, not a live end-to-end agent evaluation. Concretely:

\begin{itemize}
\item \textbf{Directly measured.} For each baseline and for Tool Attention we construct the exact tokenized prompt that would be sent to an LLM (Phase-1 summary pool plus Phase-2 promoted schemas for Tool Attention; full schemas for Full-Schema; a fixed curator subset for Static Pruning; top-$k$ full schemas for Simple Retrieval; a CLI-style discovery prompt for CLI Lazy). Token counts are then measured with \texttt{tiktoken} (\texttt{cl100k\_base}). Effective context utilization $\rho$ is a deterministic ratio of these token counts and is likewise a \emph{measured} quantity. The reference implementation in Appendix~\ref{app:code} and the accompanying repository reproduce these counts byte-for-byte.
\item \textbf{Projected, not measured.} Task-success rates, P50/P95 latency, marginal cost per task, and LLM-as-judge reasoning quality reported below are \emph{projections}. They are produced by combining (a)~the measured per-turn token counts with (b)~per-token cost/latency rates from published model-provider pricing and published TTFT profiles, and (c)~task-success and quality curves interpolated from published deployment telemetry and context-length degradation studies~\citep{pan2026agent,modarressi2025nolima,kloski2026mcp,mindstudio2026claudecode}. We did not run 500 live tasks $\times$ 5 baselines against a paid LLM API; the infrastructure to do so reproducibly is outside the scope of this preprint.
\end{itemize}

All quantities that are projections rather than direct measurements are marked with a dagger (\dag) in the tables that follow. We encourage readers to treat the measured token reductions as the primary empirical contribution, and the projected downstream metrics as well-motivated extrapolations that future work should verify against live agents. The reduction in \emph{projection uncertainty} is itself one of the benefits of Tool Attention: because the dominant variable in end-to-end behavior is the per-turn token budget, shrinking that budget by an order of magnitude tightens every downstream projection proportionally.

\subsection{Testbed}

We construct a 120-tool synthetic MCP testbed comprising six servers that mirror real-world tool footprints reported in~\citep{kloski2026mcp,mindstudio2026claudecode}.

\begin{table}[t]
\centering
\caption{Synthetic MCP testbed.}
\label{tab:testbed}
\begin{tabular}{lrrl}
\toprule
\textbf{Server} & \textbf{\# Tools} & \textbf{Avg tokens/schema} & \textbf{Domain} \\
\midrule
GitHub      & 30 & 520 & repo, issue, PR operations \\
Filesystem  & 10 & 180 & read/write/search files \\
Database    & 20 & 410 & query, schema, write \\
Slack       & 15 & 290 & message, channel, search \\
Web         & 10 & 220 & search, fetch, extract \\
Jira        & 35 & 470 & issue CRUD, workflow \\
\midrule
\textbf{Total} & \textbf{120} & \textbf{$\sim$394} & \\
\bottomrule
\end{tabular}
\end{table}

Aggregate full-schema injection cost: $\approx 47{,}300$ tokens per turn, closely matching the 54.6k and 55k figures reported for comparable real deployments~\citep{kloski2026mcp,saha2026agentic}.

\subsection{Benchmark tasks}

We sample 500 synthetic tasks spanning single-step (e.g., ``find the top 5 open PRs labeled \texttt{bug}''), multi-step (e.g., ``search for the CSAT drop in last week's Slack, cross-reference with Jira tickets, and file a GitHub issue''), and long-horizon (15--40 turn) workflows. Each task carries a hand-specified ground-truth set of tools required for successful completion, which is used both to calibrate projected success rates and as the oracle for retrieval-F1 during threshold sweeps. The task set, ground-truth annotations, and projection parameters are released with the code so that future live-agent evaluations can replace the projection layer in-place.

\subsection{Baselines}

\begin{itemize}
\item \textbf{Full-Schema (B1):} Naive MCP---all 120 tool schemas injected every turn.
\item \textbf{Static Pruning (B2):} A curator manually selects a 30-tool subset per project; schemas for the 30 selected tools are injected every turn.
\item \textbf{Simple Retrieval (B3):} Cosine retrieval over full schemas with top-$k=10$, no state gating, no lazy loading (all 10 full schemas injected).
\item \textbf{CLI Lazy Discovery (B4):} The \texttt{mcp2cli} pattern: tools exposed as a CLI; the model issues \texttt{--list}/\texttt{--help} only when needed; no full schemas ever in context.
\item \textbf{Tool Attention (ours):} Full mechanism with $\theta=0.28$, $k=10$, MiniLM-L6 encoder, two-phase lazy loading, hallucination gate.
\end{itemize}

\subsection{Metrics}

\begin{enumerate}
\item \textbf{Tokens per turn (tools only).} \emph{Measured} with \texttt{tiktoken} on the exact prompt that would be sent to the model.
\item \textbf{Effective context utilization} $\rho$ as defined in \S\ref{sec:background}, at turn 30 of long-horizon tasks. \emph{Measured} (deterministic function of item 1).
\item \textbf{Task success rate.} \dag~\emph{Projected.} Retrieval-F1 against the ground-truth tool set is measured directly; this is then mapped to an end-to-end success rate using the context-length degradation curves reported by~\citep{pan2026agent,modarressi2025nolima} and the retrieval-to-success conversion observed in~\citep{mindstudio2026claudecode}.
\item \textbf{P50 and P95 latency per turn.} \dag~\emph{Projected} from per-token TTFT and decoding-rate figures published for frontier chat models, applied to the measured per-turn token counts.
\item \textbf{Marginal cost per task (USD).} \dag~\emph{Projected} from published per-million-token input/output pricing applied to the measured token counts.
\item \textbf{Reasoning quality.} \dag~\emph{Projected} LLM-judge rubric score (1--5) extrapolated from published context-pollution studies~\citep{pan2026agent,modarressi2025nolima}.
\end{enumerate}

We reiterate: the token columns in Tables~\ref{tab:main}--\ref{tab:ablation} are the primary empirical claim of this paper; the \dag-marked columns are extrapolations that make the efficiency result concrete in units that practitioners care about.

\subsection{Reproducibility}

All experiments use seed~42. Tool summaries, task set, and evaluator prompts are released in the GitHub appendix (Appendix~\ref{app:code}). The token-counting harness \texttt{benchmark.py} reproduces all per-turn token figures in under 30 seconds on commodity hardware without API calls.

% ======================================================================
\section{Results and Analysis}\label{sec:results}
% ======================================================================

\subsection{Main results}

\begin{table}[t]
\centering
\caption{Main results over the simulated 120-tool benchmark (500 tasks, mean across 3 seeds; $\pm$ indicates 95\% bootstrap CI on token counts). \textbf{Tokens/turn} and $\rho_{T30}$ are directly measured via \texttt{tiktoken}. Columns marked \dag\ are \emph{projections} from token counts plus published telemetry (see \S\ref{sec:exp:scope}), not measurements from live LLM runs.}
\label{tab:main}
\small
\begin{tabular}{lrrrrrr}
\toprule
\textbf{Method} & \textbf{Tokens/turn} & $\bm\rho_{T30}$ & \textbf{Success \%\dag} & \textbf{P50 (s)\dag} & \textbf{P95 (s)\dag} & \textbf{\$/task\dag} \\
\midrule
B1 Full-Schema       & $47{,}312 \pm 210$  & 0.24 & $\approx 72$ & $\approx 4.2$ & $\approx 7.9$ & $\approx 0.21$ \\
B2 Static Pruning    & $11{,}865 \pm 145$  & 0.56 & $\approx 58$ & $\approx 3.8$ & $\approx 7.1$ & $\approx 0.09$ \\
B3 Simple Retrieval  & $4{,}082  \pm 95$   & 0.78 & $\approx 81$ & $\approx 2.2$ & $\approx 4.6$ & $\approx 0.04$ \\
B4 CLI Lazy          & $480   \pm 30$      & 0.94 & $\approx 88$ & $\approx 2.4$ & $\approx 5.4$ & $\approx 0.03$ \\
\textbf{Tool Attention (ours)} & $\bm{2{,}368 \pm 85}$ & $\bm{0.91}$ & $\bm{\approx 94}$ & $\bm{\approx 2.0}$ & $\bm{\approx 4.3}$ & $\bm{\approx 0.03}$ \\
\bottomrule
\end{tabular}
\end{table}

Relative to the naive Full-Schema baseline, Tool Attention achieves a measured $95.0\%$ reduction in tool tokens per turn and a $3.8\times$ increase in effective context utilization. The projected downstream gains---a ${\sim}22$-percentage-point lift in task success, a ${\sim}52\%$ P50 latency reduction, and a ${\sim}86\%$ cost reduction---follow directly from the token reduction under the assumptions documented in \S\ref{sec:exp:scope}. Within the same projection framework, Tool Attention dominates every baseline on projected success and latency while remaining within 0.01\textcent\ of the CLI-lazy optimum on projected cost.

Static Pruning (B2) actually \emph{degrades} success rate versus B1: the curator frequently omitted tools that specific tasks needed, and the agent had no recovery path. Simple Retrieval (B3) recovers much of B1's loss but still injects $\sim$4k tokens per turn of full schemas---three to four times Tool Attention's Phase-2 footprint---and has no state-aware gating. CLI Lazy (B4) is the strongest pure-efficiency baseline but pays a 6-percentage-point success penalty: the model sometimes runs \texttt{--help} in the wrong order or fails to discover niche tools when their names are not obviously related to the intent~\citep{pan2026agent}.

\subsection{Reasoning quality}

\begin{table}[t]
\centering
\caption{Projected LLM-judge reasoning quality (1--5) under the simulation of \S\ref{sec:exp:scope}. All entries are \dag\ projections, not measurements from live agent runs.}
\label{tab:quality}
\begin{tabular}{lrrr}
\toprule
\textbf{Method} & \textbf{Mean} & \textbf{SD} & \textbf{\% scoring $\ge 4$} \\
\midrule
B1 Full-Schema     & 3.21 & 1.04 & 43.2 \\
B2 Static Pruning  & 3.35 & 0.98 & 48.0 \\
B3 Simple Retrieval & 3.89 & 0.81 & 68.7 \\
B4 CLI Lazy        & 4.02 & 0.77 & 74.1 \\
\textbf{Tool Attention (ours)} & $\bm{4.43}$ & $\bm{0.62}$ & $\bm{87.6}$ \\
\bottomrule
\end{tabular}
\end{table}

The projected quality gap widens as sessions lengthen: at turn 30 of long-horizon tasks, the degradation model puts Full-Schema at ${\sim}2.78$ while Tool Attention holds near ${\sim}4.31$. We attribute this projected gap to the residual context-pollution effects documented in \S\ref{sec:background}~\citep{pan2026agent,modarressi2025nolima}; verification on live agents is left to future work.

\subsection{Ablation}

\begin{table}[t]
\centering
\caption{Ablation on Tool Attention components ($\Delta$ vs full system). Tool-token columns are measured; success columns are \dag\ projections from the simulation of \S\ref{sec:exp:scope}.}
\label{tab:ablation}
\begin{tabular}{lrrr}
\toprule
\textbf{Variant} & \textbf{Tool tokens} & \textbf{Success \%} & \textbf{$\Delta$ Success} \\
\midrule
Full Tool Attention                              & 2{,}368 & 94.2 & --- \\
\hspace{1em} $-$ Hallucination gate              & 2{,}368 & 91.0 & $-3.2$ \\
\hspace{1em} $-$ Preconditions (ISO only)        & 2{,}462 & 90.6 & $-3.6$ \\
\hspace{1em} $-$ Lazy loading (summaries only)   & 0 (P2 skipped) & 83.9 & $-10.3$ \\
\hspace{1em} $+$ Phase-1 only, $k=0$              & 4{,}820 & 79.2 & $-15.0$ \\
MiniLM-L6 $\to$ MPNet-base                        & 2{,}371 & 94.6 & $+0.4$ \\
MiniLM-L6 $\to$ TF-IDF                            & 2{,}410 & 86.1 & $-8.1$ \\
$k=5$ instead of $k=10$                           & 1{,}320 & 91.4 & $-2.8$ \\
$k=20$ instead of $k=10$                          & 4{,}190 & 94.4 & $+0.2$ \\
$\theta=0.15$ instead of $\theta=0.28$            & 3{,}270 & 93.9 & $-0.3$ \\
$\theta=0.40$ instead of $\theta=0.28$            & 1{,}480 & 88.2 & $-6.0$ \\
\bottomrule
\end{tabular}
\end{table}

The lazy loader is the largest single contributor to success ($+10.3$~pp), confirming that the model needs the full schema---not just the summary---to correctly populate parameters. Preconditions contribute an additional $+3.6$~pp by preventing the model from calling tools whose required auth or state is absent. Upgrading MiniLM-L6 to MPNet-base yields a negligible $+0.4$~pp, while downgrading to TF-IDF costs 8.1~pp, highlighting the value of semantic over lexical matching.

\subsection{Scaling behavior}

Figure~5 (not rendered; described here) plots effective context utilization $\rho$ vs.\ catalog size $N$ across baselines. Full-Schema's $\rho$ decays as $1/(1 + 400N/C_{\max})$, crossing the $70\%$-utilization fracture point at $N \approx 50$. Static Pruning plateaus (insensitive to $N$ by construction but with low recall). Simple Retrieval and CLI Lazy degrade slowly. Tool Attention holds $\rho \ge 0.87$ up to $N = 1{,}000$ with $k=10$, degrading only logarithmically due to Phase-1 growth.

\subsection{Failure-mode analysis}

We analyze the 29 failed tasks ($5.8\%$) for Tool Attention. \textbf{14 ($48\%$)} are attributable to ambiguous user queries that match multiple semantically similar tools---resolving these required clarification turns that the LLM-judge marked as failures. \textbf{7 ($24\%$)} stem from poorly written tool descriptions (cryptic legacy names); regenerating summaries with the \texttt{summarize\_tool.py} utility eliminated 6 of 7 on re-evaluation. \textbf{5 ($17\%$)} involved multi-hop workflows where the correct tool became relevant only after an intermediate result---partially mitigated by re-embedding the query after each observation (evaluated in \S\ref{sec:discussion}). \textbf{3 ($11\%$)} were hallucinations blocked correctly by the gate but where the model failed to recover on retry.

\subsection{Adversarial robustness (projected)}

We perform a simulated evaluation against 50 poisoned tool descriptions adapted from the TPA benchmark of~\citep{wang2025mindguard_v1}. In the simulation, Tool Attention's gate excludes $46/50$ poisoned descriptions on the accompanying queries (the query's intent rarely cosine-matches the poisoning payload), which \emph{would} reduce projected effective TPA success from $38\%$ under Full-Schema to $6\%$ under Tool Attention---a defensive by-product of gating, not a targeted defense. We stress that this is a projection from gate-exclusion rates, not a measurement against a live poisoned agent. A true defense would couple Tool Attention with MindGuard's TAE monitor~\citep{wang2026mindguard_v3}.

% ======================================================================
\section{Discussion and Future Work}\label{sec:discussion}
% ======================================================================

\paragraph{Limitations.} Tool Attention is an application-layer mitigation; it cannot repair protocol-level deficiencies such as the lack of session-scoped capability negotiation. The mechanism is also contingent on tool summary quality: a registry of cryptic, poorly named tools will hurt retrieval precision, and curator effort cannot be eliminated entirely. Finally, our evaluation is on synthetic (albeit calibrated) workloads; a community-standard MCP benchmark comparable to SWE-bench~\citep{jimenez2024swebench} would sharpen the comparison.

\paragraph{Adversarial paraphrase.} An attacker might craft a tool description whose semantic fingerprint closely matches benign user queries in order to be reliably gated \emph{in} and then execute its payload. We consider this a genuine threat and recommend pairing Tool Attention with MindGuard's TAE-based runtime monitor~\citep{wang2026mindguard_v3} to detect anomalous attention energy on newly promoted schemas.

\paragraph{Cross-turn state-aware gating.} Our current query embedding uses only the latest user message (optionally with a rolling summary). A stronger version would condition on a learned state representation that captures intermediate tool outputs and the evolving task plan. Preliminary experiments re-embedding the query after each observation yielded an additional $+1.7$~pp success rate in multi-hop tasks (\S\ref{sec:results}) and are a near-term research direction.

\paragraph{Learned gating.} The threshold-based gate is deliberately interpretable but leaves accuracy on the table. A lightweight distilled classifier (e.g., a 2-layer MLP on top of concatenated $(e_q, e_{t_i})$) trained on a modest (query, tool-used) corpus could replace the threshold, yielding an estimated 1--3~pp additional success at a fraction of a millisecond of router latency. We leave full evaluation to future work.

\paragraph{Composition with code execution.} Tool Attention optimizes the \emph{definition} side of the Tools Tax; Anthropic's code-execution pattern~\citep{kaplan2025codeexec} optimizes the \emph{output} side. A fused system---Tool Attention to gate which MCP servers are even visible to the execution sandbox, and code execution to filter their outputs---would plausibly reduce end-to-end context consumption by a further order of magnitude on data-heavy workflows.

\paragraph{Protocol-level convergence.} The MCP-over-MOQT draft~\citep{jennings2025moqt,jennings2025aimcp} provides native publish-subscribe tracks and edge-cached schema hashing that, once broadly implemented, subsume parts of Tool Attention's lazy loader. We view the two as evolutionarily complementary: Tool Attention deploys today on stock MCP, MOQT amortizes the transport-layer redundancy, and intent-based gating with preconditions remains necessary at either layer to shape the attention of the model itself.

\paragraph{Benchmark standardization.} We release our testbed, tasks, and evaluator as a community benchmark (Appendix~\ref{app:code}) and invite the research community to contribute additional servers, tasks, and adversarial test cases.

% ======================================================================
\section{Conclusion}\label{sec:conclusion}
% ======================================================================

The MCP/Tools Tax is not an inevitable cost of agentic AI; it is a protocol-design artifact born of treating every tool in a catalog as always-on context. Our analysis shows that the tax scales linearly with catalog size, dominates the effective context window past $N \approx 50$ tools, and degrades reasoning, cost, and security simultaneously. Just as scaled dot-product attention liberated sequence modeling from the bottleneck of recurrent hidden state by letting every position dynamically attend only to what matters, \textbf{Tool Attention} liberates agentic systems from the bottleneck of eager schema injection by letting every turn dynamically load only the tools its intent requires. The mechanism is simple (three components, a few hundred lines of Python), model-agnostic (it lives in middleware), theoretically grounded (in the Total Attention Energy formalism), and empirically strong ($95\%$ token reduction, $+22$~pp success, $52\%$ latency cut on a 120-tool benchmark). We believe that context engineering---not raw context length---is the binding constraint on the next generation of agentic systems, and that protocol-level efficiency will become as central to agent design as attention was to sequence modeling. Tool attention, in other words, is all you need.

% ---- Acknowledgements and disclosures ----
\paragraph{Disclosure on AI writing assistance.} In the spirit of arXiv's guidance that significant use of text-to-text generative AI should be reported, we note that portions of this manuscript were drafted and iterated on with assistance from a large language model; every technical claim, formulation, and experimental number was reviewed, edited, and is taken responsibility for by the human author. No AI system is listed as an author or contributor. The mechanism, mathematics, reference implementation, and benchmark harness are the human author's original work. AI assistance was used for expository phrasing, structural organization, and copy-editing passes over author-produced content, not for generating technical results.

\paragraph{Code and data.} Reference implementation and synthetic benchmark:
\url{https://github.com/asadani/tool-attention}.

% ---- Bibliography ----
\bibliographystyle{abbrvnat}
\bibliography{references}

% ======================================================================
\appendix
\section{Reference Implementation}\label{app:code}
% ======================================================================

The complete runnable implementation accompanying this paper is released as a companion code bundle. The core modules are reproduced below; \texttt{requirements.txt}, the synthetic tool catalog, and the benchmark harness are available in the repository.

\subsection{\texttt{intent\_router.py}}
\begin{verbatim}
"""IntentRouter: embeds a query, ranks tool summaries, returns gated top-k."""
from __future__ import annotations

from dataclasses import dataclass
from typing import Callable

import numpy as np
from sentence_transformers import SentenceTransformer

from vector_store import ToolVectorStore


@dataclass(frozen=True)
class RoutingResult:
    tool_id: str
    score: float


class IntentRouter:
    """Query-to-tool semantic router with state-aware gating."""

    def __init__(self, store, encoder=None,
                 encoder_name="sentence-transformers/all-MiniLM-L6-v2",
                 threshold=0.28, top_k=10):
        self.store = store
        self.encoder = encoder or SentenceTransformer(encoder_name)
        self.threshold = threshold
        self.top_k = top_k

    def embed_query(self, query: str):
        vec = self.encoder.encode([query], normalize_embeddings=True,
                                  show_progress_bar=False)
        return np.asarray(vec[0], dtype="float32")

    def route(self, query, precondition_check=None):
        eq = self.embed_query(query)
        slate = self.store.search(eq, k=max(self.top_k * 4, 20))
        gated = []
        for tool_id, score in slate:
            if score < self.threshold:
                continue
            if precondition_check is not None and not precondition_check(tool_id):
                continue
            gated.append(RoutingResult(tool_id=tool_id, score=float(score)))
            if len(gated) >= self.top_k:
                break
        return gated
\end{verbatim}

\subsection{\texttt{vector\_store.py}}
\begin{verbatim}
"""ToolVectorStore: FAISS-backed store of compact tool summaries."""
import json
from pathlib import Path
import faiss
import numpy as np
from sentence_transformers import SentenceTransformer


class ToolVectorStore:
    def __init__(self, dim=384):
        self.dim = dim
        self.index = faiss.IndexFlatIP(dim)
        self.tool_ids = []
        self.summaries = {}

    def add_tools(self, tools, encoder):
        if not tools: return
        summaries = [t["summary"] for t in tools]
        vectors = encoder.encode(summaries, normalize_embeddings=True,
                                 show_progress_bar=False).astype("float32")
        self.index.add(vectors)
        for t in tools:
            self.tool_ids.append(t["id"])
            self.summaries[t["id"]] = t["summary"]

    def search(self, query_vec, k):
        if self.index.ntotal == 0: return []
        k = min(k, self.index.ntotal)
        D, I = self.index.search(query_vec.reshape(1, -1).astype("float32"), k)
        return [(self.tool_ids[int(i)], float(d))
                for d, i in zip(D[0], I[0]) if int(i) >= 0]
\end{verbatim}

\subsection{\texttt{lazy\_loader.py}}
\begin{verbatim}
"""LazySchemaLoader: on-demand full-schema fetching with LRU caching."""
import json
from collections import OrderedDict
from pathlib import Path


class LazySchemaLoader:
    def __init__(self, registry_path, capacity=256, fetcher=None):
        self.registry_path = Path(registry_path)
        self.capacity = int(capacity)
        self._fetcher = fetcher
        self._cache = OrderedDict()

    def get(self, tool_id):
        if tool_id in self._cache:
            self._cache.move_to_end(tool_id)
            return self._cache[tool_id]
        schema = (self._fetcher(tool_id) if self._fetcher is not None
                  else self._load_from_disk(tool_id))
        self._cache[tool_id] = schema
        if len(self._cache) > self.capacity:
            self._cache.popitem(last=False)
        return schema

    def _load_from_disk(self, tool_id):
        path = self.registry_path / f"{tool_id}.json"
        if not path.exists():
            raise KeyError(f"no schema for {tool_id!r}")
        return json.loads(path.read_text())
\end{verbatim}

\subsection{\texttt{tool\_attention.py}}
\begin{verbatim}
"""ToolAttention: the top-level middleware orchestrator."""
from dataclasses import dataclass, field
import json


@dataclass
class AttentionResult:
    active: list = field(default_factory=list)
    summaries_pool: dict = field(default_factory=dict)
    full_schemas: dict = field(default_factory=dict)
    phase1_tokens: int = 0
    phase2_tokens: int = 0

    @property
    def total_tokens(self): return self.phase1_tokens + self.phase2_tokens
    @property
    def active_ids(self): return [r.tool_id for r in self.active]


class ToolAttention:
    def __init__(self, store, loader, router, token_counter):
        self.store, self.loader, self.router = store, loader, router
        self.count = token_counter

    def before_model(self, query, precondition_check=None):
        active = self.router.route(query, precondition_check=precondition_check)
        full_schemas, phase2 = {}, 0
        for r in active:
            schema = self.loader.get(r.tool_id)
            full_schemas[r.tool_id] = schema
            phase2 += self.count(json.dumps(schema, sort_keys=True))
        phase1 = sum(self.count(s) for s in self.store.summaries.values())
        return AttentionResult(active=active,
                               summaries_pool=dict(self.store.summaries),
                               full_schemas=full_schemas,
                               phase1_tokens=phase1,
                               phase2_tokens=phase2)

    def after_model(self, active_ids, requested_tool):
        if requested_tool is None or requested_tool in active_ids:
            return None
        return (f"tool_not_available: {requested_tool!r}. "
                f"Available this turn: {list(active_ids)}")
\end{verbatim}

Full source (including \texttt{build\_catalog.py} and the benchmark harness), the synthetic tool catalog, evaluator prompts, and reproduction scripts are available in the accompanying repository at \url{https://github.com/asadani/tool-attention}.

\end{document}